\pdfoutput=1
\documentclass[11pt]{article}

\usepackage[]{emnlp2021}

\usepackage{times}
\usepackage{latexsym}

\usepackage[T1]{fontenc}

\usepackage[utf8]{inputenc}

\usepackage{microtype}

\usepackage[finalnew]{trackchanges}
\addeditor{YJ}
\addeditor{WD}
\newcommand{\newparagraph}[1]{\noindent\textbf{#1}}

\title{\textsc{SideControl}: Controlled Open-domain Dialogue Generation via Additive Side Networks}

\author{Wanyu Du\quad Yangfeng Ji \\
    Department of Computer Science \\
    University of Virginia \\
    Charlottesville, VA 22904 \\
    \texttt{\{wd5jq,yangfeng\}@virginia.edu} \\
}

\usepackage{amsmath,amsfonts,amssymb,bm}
\usepackage{pifont} 
\usepackage{graphicx,epsfig,fancybox} 
\usepackage{float}
\usepackage{color} 
\usepackage[ruled,vlined]{algorithm2e} %
\usepackage{algorithmic}
\usepackage{multirow}
\usepackage{comment}
\usepackage{hyperref}
\usepackage{natbib} 
\usepackage{subfigure}

\renewcommand{\vec}[1]{\boldsymbol{#1}}

\newcommand{\mat}[1]{\mathbf{#1}}

\usepackage{adjustbox}
\usepackage{array}
\usepackage{booktabs}

\newcolumntype{R}[2]{%
    >{\adjustbox{angle=#1,lap=\width-(#2)}\bgroup}%
    l%
    <{\egroup}%
}

\begin{document}
\maketitle

\begin{abstract}
Transformer-based pre-trained language models boost the performance of open-domain dialogue systems.
Prior works leverage Transformer-based pre-trained language models to generate texts with desired attributes in two general approaches: 
(1) gradient-based methods: updating all latent representations of pre-trained models with gradients from attribute models;
(2) weighted-decoding methods: re-ranking beam candidates from pre-trained models with attribute functions. 
However, gradient-based methods lead to high computation cost and can easily get overfitted on small training sets, while weighted-decoding methods are inherently constrained by the low-variance high-bias pre-trained model.
In this work, we propose a novel approach to control the generation of Transformer-based pre-trained language models: the \textsc{SideControl} framework, 
\change[WD]{which learns a small amount of side (task-specific) parameters guided by the control attributes loss,  
in order to incorporate useful control signals into the final word probability distribution while maintaining a relatively low computation cost.} 
{which leverages a novel control attributes loss to incorporate useful control signals, and is shown to perform well with very limited training samples.}
\change[WD]{We evaluate the controllability and generation quality of our proposed method on two benchmark open-domain dialogue datasets, and results show that the \textsc{SideControl} framework is 6x-24x faster than gradient-based baseline during generation, and outperforms weighted-decoding baseline in controllability around 12\%.}
{We evaluate our proposed method on two benchmark open-domain dialogue datasets, and results show that the \textsc{SideControl} framework has better controllability, higher generation quality and better sample-efficiency than existing gradient-based and weighted-decoding baselines.} 
\footnote{Our code implementation and data sources can be found here: \url{https://github.com/wyu-du/Controlled-Dialogue-Generation}.}
\end{abstract}

\section{Introduction}


With the advance of Transformer-based pre-trained language models \citep{radford2019language,2020t5,NEURIPS2020_1457c0d6,zhang-etal-2020-dialogpt}, many dialogue systems \citep{zhang-etal-2020-dialogpt,roller2020recipes, shuster-etal-2020-dialogue} have shown promising performance in challenging open-domain conversations with humans.
However, for controlled dialogue generation, prior works mainly focus on building LSTM-based class-conditional generative model on specific datasets with task-specific design on model architecture \citep{wen-etal-2015-semantically,ke-etal-2018-generating,chen-etal-2019-semantically,see-etal-2019-makes} or policy learning strategy \citep{kawano-etal-2019-neural,hsueh-ma-2020-semantic,takayama-arase-2020-consistent,varshney-etal-2021-modelling}.
In this work, we explore effective method for controlled generation on Transformer-based dialogue systems, with the goal of adding controllability functionality into state-of-the-art Transformer-based dialogue systems with lower computation cost, less training data and more flexible control mechanism.

\begin{figure*}[t]
  \centering
  \subfigure[SideNet for Knowledge Document Control]{
    \includegraphics[width=0.5\linewidth]{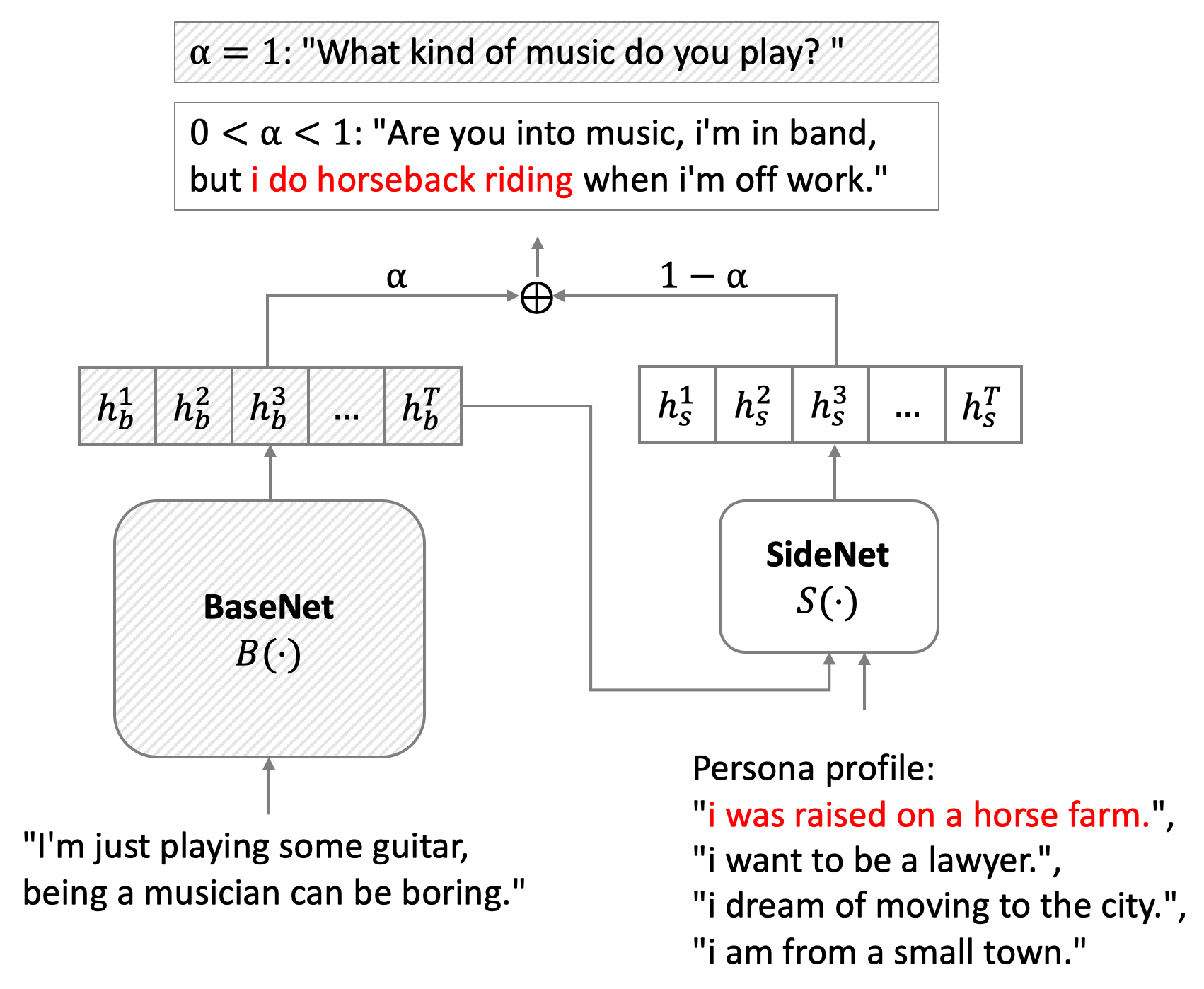}
  }
  \subfigure[SideNet for Semantic Label Control]{
    \includegraphics[width=0.47\linewidth]{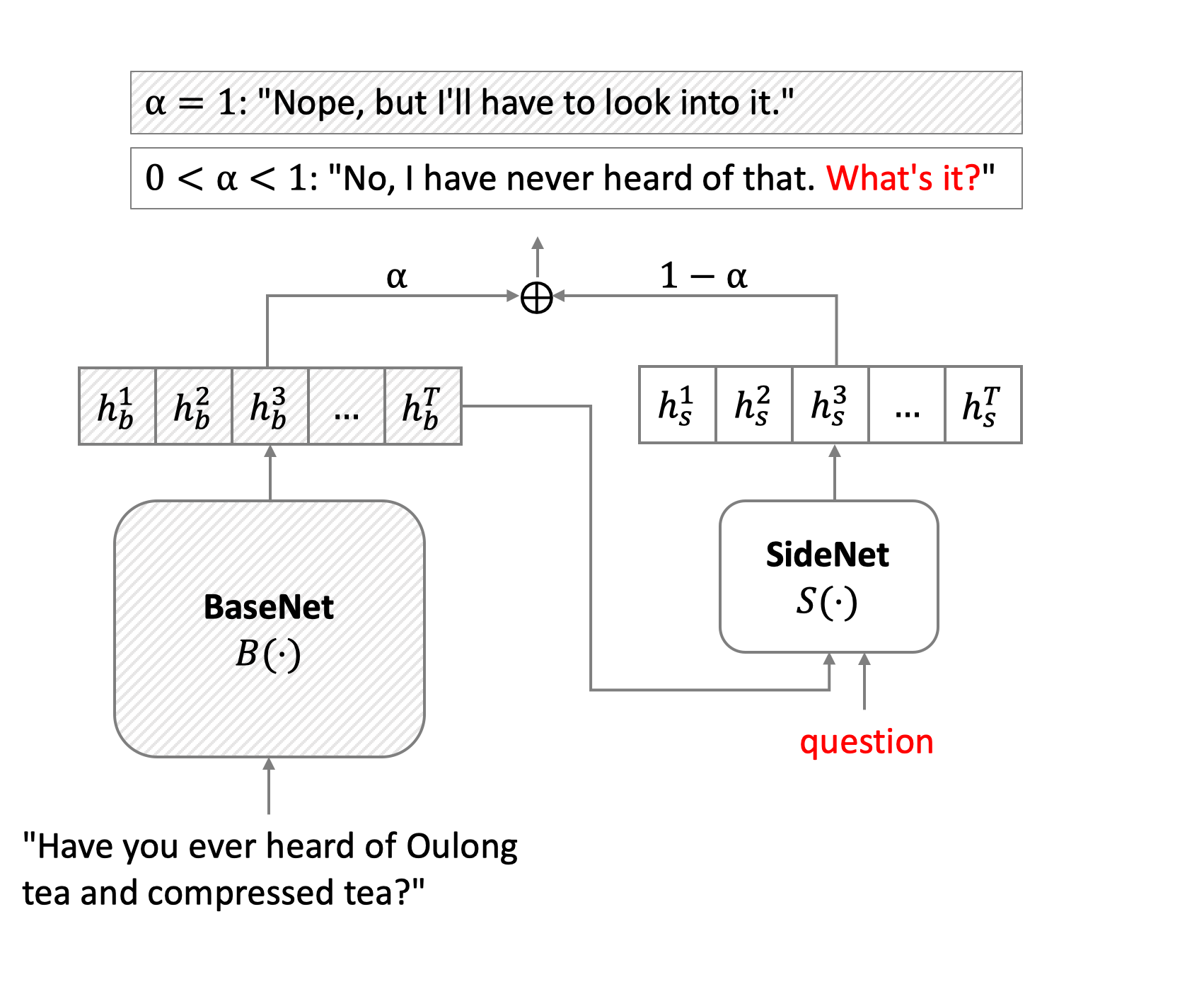}
  }
  \caption{
  General architecture of the \textsc{SideControl} framework.
    \label{fig:model-arch}}
\end{figure*}

Prior works on controlled text generation for Transformer-based pre-trained language models can be categorized into two general approaches: (1) gradient-based methods and (2) weighted-decoding methods.
The gradient-based methods \citep{DBLP:journals/corr/abs-1912-02164,goswamy-etal-2020-adapting,DBLP:journals/corr/abs-2104-04039} propose a plug-and-play language model following $p(x|a)\propto p(a|x)p(x)$, which plugs an attribute model $p(a|x)$ with a pre-trained language model $p(x)$ to control generation.
The gradients from $p(a|x)$ are used to guide the latent representations of pre-trained models encoding more control attribute information.
The weighted-decoding methods \citep{ghazvininejad-etal-2017-hafez,baheti-etal-2018-generating,holtzman-etal-2018-learning,yang-klein-2021-fudge} modify the sampling weights with attribute functions in beam search at each decoding timestep to control generation.
Essentially, the attribute functions are used to re-rank the original beam candidates generated by the pre-trained language models.
{The main idea of both gradient-based methods and weighted decoding methods is the flexibility: users can design any attribute models or functions for different controlled generation tasks and apply the attribute model or function to any state-of-the-art pre-trained language models for generating high quality texts.}

However, weighted decoding methods \citep{ghazvininejad-etal-2017-hafez,baheti-etal-2018-generating,holtzman-etal-2018-learning,yang-klein-2021-fudge} are limited by {the low-variance high-biased pre-trained language models}, 
since they do not update the pre-trained language models.
If the pre-trained model yields commonly observed words rather than target attribute words in the beam candidates list, it is difficult for the attribute functions to re-rank and find the target words during generation.
Although gradient-based methods \citep{DBLP:journals/corr/abs-1912-02164,goswamy-etal-2020-adapting,DBLP:journals/corr/abs-2104-04039} do not have this limitation since they update the latent representations of pre-trained models during inference, the gradient propagation at each decoding timestep involves heavy computation, which results in slow response speed to users.
In addition, the controllability performance of gradient-based methods relies on the attribute model.
If the attribute model gets overfitted on a small training set, the gradient from this attribute model will just lead to meaningless updates.

To build an effective and efficient controlled open-domain dialogue system, we propose the \textsc{SideControl} framework, which treats the pre-trained lanaguage model as a feature extractor and train light-weight side networks to encode complementary information from control attributes.
In addition, we introduce a novel control attributes loss to guide the side network during training.
As shown in \autoref{fig:model-arch}, the final output representation is a mixture of a base representation from the pre-trained language model and a side representation from the side network.
The mixture coefficient $\alpha$ is learned during training, and is used to balance the prior knowledge from the base network and the task-specific control attributes signals from the side network.
From the encoding perspective, the \textsc{SideControl} framework not only can be applied to any pre-trained language models, but also supports diverse format attributes control (e.g. dialogue act, external knowledge document).
From the decoding perspective, the \textsc{SideControl} framework has low computation cost, since it directly samples from its optimized class-conditional language model $p(x|a)$ without additionally updating latent representations during generation.
From the sample-efficiency perspective, the \textsc{SideControl} framework achieves good performance with a few thousand training samples by leveraging the control loss.

We summarize the contributions of this work as follows:
\begin{enumerate}
\item we propose a new controlled dialogue generation framework with novel control attributes losses to support different forms of attributes control (e.g. dialogue act, external knowledge document);
\item 
\change[WD]{we conduct empirical experiments to show that the \textsc{SideControl} framework is effective even with $100\sim1000$ training samples;}
{we conduct empirical experiments to show the sample-efficiency of the \textsc{SideControl} framework, which can achieve good performance with only $100\sim1000$ training samples;}
\item we conduct empirical experiments to validate that the \textsc{SideControl} framework has better controllability, better text quality, and lower \change[WD]{computation}{decoding} cost compared to gradient-based methods and weighted-decoding methods.
\end{enumerate}

\section{SideNet for Controlled Generation}

Firstly, we introduce the \textsc{SideControl} framework in \autoref{sec:general}, which presents the general idea of using a small side network to coordinate the generation process based on large-scale pre-trained language models \citep{zhang-etal-2020-dialogpt,roller2020recipes, shuster-etal-2020-dialogue}.
Then we provide two realizations of side networks for two types of control attributes: (1) external knowledge document in \autoref{sec:continuous-control}, (2) semantic label in \autoref{sec:discrete-control}.

\subsection{General Framework}
\label{sec:general}
Given a dialogue context which contains a fixed number of previous utterances $\mat{X}=\{\vec{x}_i\}_{i=1}^N$, where $N$ is the total number of tokens in the given dialogue context, 
and a control attribute $\vec{a}$ which represents the desired controllable attributes, 
the goal is to build a model conditioned on $\mat{X}$ and $\vec{a}$ that can generate a response which best approximates the ground-truth human response $\mat{Y}=\{\vec{y}_t\}_{t=1}^T$:
\begin{eqnarray}
  p(\mat{Y}\mid \mat{X}, \vec{a})
  &=& \prod_{t=1}^T p(\vec{y}_t\mid \vec{y}_{1:t-1}, \vec{x}_{1:N}, \vec{a})\nonumber\\
  &=& \prod_{t=1}^T p(\vec{y}_t\mid \vec{h}_t)
\end{eqnarray}
where $\vec{h}_t$ is the last hidden state of the generative model at decoding timestep $t$.

The \textsc{SideControl} framework consists of a large base network $B(\cdot)$  providing rich feature representations and a small side network $S(\cdot)$ encoding control attribute(s), as illustrated in \autoref{fig:model-arch}. 
The base network $B(\cdot)$ can be any pre-trained language models \citep{zhang-etal-2020-dialogpt,roller2020recipes, shuster-etal-2020-dialogue}.
Given dialogue context $\vec{x}_{1:N}$ as the input to the base network, we just take last hidden states $\{\vec{h}_b^t\}_{t=1}^T$ for the response $\{\vec{y}_t\}_{t=1}^T$ from the base network as our base representations:  
\begin{equation}
  \vec{h}_b^1, \dots, \vec{h}_b^T = B(\vec{x}_{1:N})
\end{equation}
The side network $S(\cdot)$ is a light-weight neural network, which encodes the control attribute $\vec{a}$ into base representations $\{\vec{h}_b^t\}_{t=1}^T$:
\begin{equation}
    \vec{h}_s^1, \dots, \vec{h}_s^T = S(\vec{a}, \vec{h}_b^1, \dots, \vec{h}_b^T)
\end{equation}
Finally, we keep the base representation $\vec{h}_b^t$ fixed, and add the side representation $\vec{h}_s^t$ upon it to obtain the final combined representation $\vec{h}_t$ for the current token $\vec{y}_t$:
\begin{eqnarray}
  \label{eq:mixture}
  \vec{h}_t &=& \alpha \cdot \vec{h}_b^t + (1-\alpha) \cdot \vec{h}_{s}^t  \\
  p(\vec{y}_t\mid\vec{h}_t) &=& \text{softmax}(\mat{W}_{\text{vocab}}\vec{h}_t) \label{eq:global-alpha}
\end{eqnarray}
where $\mat{W}_{\text{vocab}}$ is learnable parameters, and the mixture coefficient $\alpha$ is also learned during training, which aims to encode both useful prior knowledge from pre-trained language models and important attribute information from target dataset for controlled generation. 
We provide detailed implementations for the side network $S(\cdot)$ and mixture coefficient $\alpha$ in \autoref{sec:discrete-control} and \autoref{sec:continuous-control}.

The \emph{main challenge} in this framework is to teach the side network $S(\cdot)$, such that it can provide complementary information of control signals via $\vec{h}^{t}_s$ during generation, since the pre-trained language models can already generate fluent responses.
To address this challenge, we intentionally freeze the parameters of the base network $B(\cdot)$ when training the side network.
Otherwise, it is essentially training a large neural network model even deeper than $B(\cdot)$.
Second, we introduce the control attribute loss $\mathcal{L}_{control}$, which is designed to teach the side network explicitly encoding control signals to improve the controllability of the model.
The final objective is a combination of class-conditional language modelling loss $\mathcal{L}_{cclm}$ and task-specific control attributes loss $\mathcal{L}_{control}$:
\begin{equation}
    \mathcal{L} = \mathcal{L}_{cclm} + \lambda\cdot \mathcal{L}_{control} \label{eq:general-objective} 
\end{equation}
where 
$\lambda$ is a task-specific hyper-parameter,
and detailed implementations of $\mathcal{L}_{cclm}$ and $\mathcal{L}_{control}$ are described in \autoref{sec:continuous-control} and \autoref{sec:discrete-control}, $\mathcal{L}_{control}$ has different implementation when controlling different forms of attributes.

\subsection{Knowledge Document Control}
\label{sec:continuous-control}
When having external knowledge documents as the control attributes, such as persona profile \citep{dinan2020second}, Wikipedia articles \citep{dinan2018wizard}, etc., the format of control attribute is sequences of tokens $\vec{a}=\{\vec{k}_i\}_{i=1}^K$, where $K$ is the total number of tokens in the external knowledge document.
In this case, we model the knowledge document representation with a single-layer bi-directional LSTM:
\begin{equation}
    \vec{h}_k^1, ..., \vec{h}_k^K = \text{BiLSTM}(\vec{a})
\end{equation}
The side network is designed to align the controlled knowledge document representation $\{\vec{h}_k^i\}_{i=1}^K$ with the base representation $\vec{h}_b^t$ at each decoding timestep.
We compute the cross-attention between $\{\vec{h}_k^i\}_{i=1}^K$ and $\vec{h}_b^t$ following \citep{bahdanau2014neural}:
\begin{eqnarray}
    e_i^t &=& v^T\cdot\tanh{(\mat{W}_k \vec{h}_k^i + \mat{W}_b \vec{h}_b^t + b_{kb})} \\
    a_i^t &=& \text{softmax}(e_i^t) \\
    \vec{c}_k^t &=& \sum_{i=1}^K a_i^t \cdot \vec{h}_k^i
\end{eqnarray}
where $\mat{W}_k \in \mathbb{R}^{D\times D}$, $\mat{W}_b \in \mathbb{R}^{D\times D}$ and $b_{kb} \in \mathbb{R}^{D}$ are learnable parameters.
The attention $a^t$ is a probability distribution over the controlled knowledge document that tells the decoder where to look at when generating the next word, and the context vector $\vec{c}_k^t$ represents what has been read from the controlled knowledge document representation at decoding timestep $t$.
The final side representation $\vec{h}_s^t$ incorporates the context vector $\vec{c}_k^t$ into the base representation $\vec{h}_b^t$:
\begin{equation}
    \vec{h}_s^t = \tanh{(\mat{W}_c[\vec{c}_k^t; \vec{h}_b^t]+b_c)}
\end{equation}
where we concatenate $\vec{c}_k^t$ and $\vec{h}_b^t$, and $\mat{W}_c \in \mathbb{R}^{2D\times D}$ and $b_c \in \mathbb{R}^D$ are learnable parameters.

Since the controlled knowledge document is different per utterance, we implement the mixture coefficient $\alpha$ based on the side representation $\vec{h}_s^t$ and base representation $\vec{h}_b^t$ at decoding timestep $t$:
\begin{eqnarray}
    \alpha_t &=& \sigma(\mat{W}_\alpha[\vec{h}_s^t; \vec{h}_b^t]+b_\alpha)\\
    \vec{h}_t &=& \alpha_t \cdot \vec{h}_b^t + (1-\alpha_t) \cdot \vec{h}_{s}^t \label{eq:kb-ht}
\end{eqnarray}
where we concatenate $\vec{h}_s^t$ and $\vec{h}_b^t$, and $\mat{W}_\alpha \in \mathbb{R}^{2D\times 1}$ and $b_\alpha \in \mathbb{R}$ are learnable parameters.

In order to encourage the decoder generating more words from the knowledge document, we adopt the copy mechanism from \citep{see-etal-2017-get} to formulate $\mathcal{L}_{cclm}$:
\begin{eqnarray}
  \beta &=& \sigma(\mat{W}_\beta[\vec{c}_k^t; \vec{h}_b^t]+b_\beta) \\
  p(\vec{y}_t\mid \vec{h}_t) &=& \beta p(\vec{y}_t|\vec{h}_t) + (1-\beta)\sum_{i=1}^K a_i^t \\
  \mathcal{L}_{cclm} &=& -\sum_{t=1}^T \log p(\vec{y}_t^*\mid \vec{h}_t)
\end{eqnarray}
where we concatenate $\vec{c}_k^t$ and $\vec{h}_b^t$, and $\mat{W}_\beta \in \mathbb{R}^{2D\times 1}$ and $b_\beta \in \mathbb{R}$ are learnable parameters.
$\vec{h}_t$ comes from \autoref{eq:kb-ht}.
$\vec{y}_t^*$ is the ground-truth word at decoding timestep $t$.
$\sum_{i=1}^K a_i^t$ is the summation of attention distribution over the knowledge document at current decoding timestep $t$, which will assign higher probability for attended knowledge document words in the final word probability distribution.

The control attributes loss for this task is used to encourage generating more non-repetitive words from the knowledge document.
We adopt the {coverage mechanism} from \citep{see-etal-2017-get} to formulate $\mathcal{L}_{control}$:
\begin{equation}
    \mathcal{L}_{control} = \sum_{t=1}^T \sum_{i=1}^K \min(a_i^t, \sum_{t'=0}^{t-1}a_i^{t'}) \label{eq:kb-control-loss}
\end{equation}
where $a_i^{t'}$ is the attention weight of knowledge document word $\vec{k}_i$ at previous decoding time step $t'$.
$\mathcal{L}_{control}$ penalizes the overlap between current attention distribution and previous attention distributions, which prevents the model repeatedly attending to the same word in the knowledge document.
For more details about the copy mechanism and coverage mechanism, please refer to the original paper \citep{see-etal-2017-get}.


\subsection{Semantic Label Control}
\label{sec:discrete-control}
When having a semantic label as the control attribute,  
such as dialogue act \citep{li-etal-2017-dailydialog}, emotion \citep{rashkin-etal-2019-towards}, etc., we implement the side network as a simple feed-forward neural network:
\begin{eqnarray}
    \vec{h}_{s}^t &=& \tanh{(\mat{W}_d[\mat{W}_a\vec{a}; \vec{h}_{b}^t] + b_d)} \\
    \vec{h}_t &=& \alpha\cdot \vec{h}_b^t + (1-\alpha)\cdot \vec{h}_s^t \\
    \mathcal{L}_{cclm} &=& -\sum_{t=1}^T \log p(\vec{y}^*_t\mid \vec{h}_t)
\end{eqnarray}
where we concatenate $\mat{W}_a\vec{a}$ and $\vec{h}_b^t$, $\mat{W}_a \in \mathbb{R}^{1\times D}$ is an embedding matrix that maps the discrete label $\vec{a}$ to a continuous representation, $\mat{W}_d \in \mathbb{R}^{2D \times D}$ and $b_d \in \mathbb{R}^{D}$ are learnable parameters.
The mixture coefficient $\alpha\in [0,1]$ is a global parameter which is learned during training, in order to encode both useful prior knowledge from pre-trained language models and control signals from semantic label.
$\vec{y}^*_t$ is the ground-truth word at decoding timestep $t$.

The control attributes loss {$\mathcal{L}_{control}$} for this task is used to modify the final latent representations so that the model can generate responses with the target control attribute.
However, it is difficult to directly measure how much control attribute information has been encoded into the side representation. 
Therefore, we approximate it using a independent attribute classifier $p(\vec{a}|\vec{h}_{1:T})$.
When training the side network, we keep the attribute classifier fixed and feed the side representations $\{\vec{h}_s^t\}_{t=1}^T$ into the classifier.
The classifier will return a loss between the current side representation and the target control attribute $\vec{a}^*$, and optimizing this loss will update the side representation $\vec{h}_s^t$ towards obtaining a higher $p(\vec{a}^*|\vec{h}_{1:T})$:
\begin{eqnarray}
    p(\vec{a}\mid \vec{h}_{1:T}) &=& \text{softmax}(\mat{W}_{\text{clf}} \frac{\sum_{t=1}^T\vec{h}_s^t}{T}) \\
    \mathcal{L}_{control} &=& -\log p(\vec{a}^*\mid \vec{h}_{1:T}) \label{eq:da-control-loss}
\end{eqnarray}
Note that $\mat{W}_{\text{clf}} \in \mathbb{R}^{D\times K}$ is independently learned on the same training set based on the base representation $\{\vec{h}_b^t\}_{t=1}^T$, but is fixed when we update the side network.

\section{Experiments}

\subsection{Evaluation Methods}

In this work, we focus on evaluating the controllability and text quality of different controlled generation methods.
Additionally, we prefer to have lower decoding cost and better modularity in order to apply the proposed method into more possible applications.
Therefore, we use the following automatic metrics to evaluate the performance:\\
\newparagraph{Controllability}\footnote{We provide implementation details in \autoref{sec:metrics-control}}: this is our main metric. It aims at evaluating whether the proposed method can successfully generate the target controlling attributes. 
\begin{enumerate}
\item For the semantic label control task, we use the \textit{classification accuracy} computed by an independently trained BERT classifier \citep{devlin-etal-2019-bert}. 
\item For the knowledge document control task, we use the \textit{cosine similarity} between the word vectors of external knowledge document and model generated response. We use the pre-trained GloVe embedding \citep{pennington-etal-2014-glove} to model the word vectors.
\end{enumerate}
    
\newparagraph{Text Quality}: it aims at evaluating how well the model learns to generate responses that match the ground-truth references, 
where we use model perplexity (PPL) computed on the test set \footnote{Note that PPLM and FUDGE do not update the generative model and are applied only during generation, which means their model perplexity will be the same with their base network, i.e. DialoGPT-Ori, therefore we do not report their model perplexity in performance results.}, BLEU \citep{papineni2002bleu} and METEOR \citep{banerjee-lavie-2005-meteor} to approximate it.

\newparagraph{Decoding Cost}: it evaluates the generation efficiency of the proposed method.
Given the same set of 10 dialogue contexts, we compute the decoding time per token across different methods, a faster decoding time indicates the method is more efficient during generation.

\newparagraph{Modularity}: it evaluate how well the side network can be applied to different base networks. We compare model performance under two different types of pre-trained language models: DialoGPT \citep{zhang-etal-2020-dialogpt} and BlenderBot \citep{roller2020recipes}.
Ideally, we expect as good or even better performance when switching the base network from DialoGPT to BlenderBot, since BlenderBot has been trained on larger dialogue corpus that is likely to provide more informative base representations.

\subsection{Competitive Baselines}
We compare the \textsc{SideControl} framework with the following competitive baselines:\\
\newparagraph{DialoGPT-Ori}: the original pre-trained language model for open-domain dialogue generation, DialoGPT \citep{zhang-etal-2020-dialogpt} 
DialoGPT is a Transformer-based language model. It is the baseline for all other controlled generation methods.

\newparagraph{DialoGPT-FT}: direct fine-tuning the DialoGPT on the target dialogue dataset. It is used as a strong baseline for evaluating the generation quality of the generative model.

\newparagraph{DialoGPT-PPLM}: the Plug-and-Play Language Model (PPLM) \citep{DBLP:journals/corr/abs-1912-02164} with DialoGPT as the base pre-trained language model. It is a strong gradient-based baseline.

\newparagraph{DialoGPT-FUDGE}: the Future Discriminators for Generation (FUDGE) \citep{yang-klein-2021-fudge} with DialoGPT as the base pre-trained language model. It is a strong weighted decoding baseline.

\newparagraph{DialoGPT-SideControl}: our \textsc{SideControl} framework with DialoGPT as the base pre-trained language model. It is used to validate the effectiveness of our side network compared with other controlled generation baselines.

\begin{table*}[h!]
  \centering
  \small
  \begin{tabular}{lcccccc}
    \toprule
    & \multicolumn{1}{c}{\textbf{Controllability}} & \multicolumn{4}{c}{\textbf{Text Quality}} & \multicolumn{1}{c}{\textbf{Decoding Cost}} \\
     \cmidrule(lr){2-2} \cmidrule(lr){3-6} \cmidrule(lr){7-7}
    \textsc{Method} & \textsc{Similarity} $\uparrow$ & \textsc{PPL} $\downarrow$ & \textsc{BLEU-1} $\uparrow$  & \textsc{BLEU-2} $\uparrow$  & \textsc{METEOR} $\uparrow$ & \textsc{Time} $\downarrow$  \\
    \midrule
    DialoGPT-Ori & 0.6382 & 68.63 & 12.95 & 1.22 & 0.0526 & 0.0786 s/tok \\
    DialoGPT-FT & 0.6732 & 15.22 & 17.27 & 2.05 & 0.0675 & 0.0721 s/tok \\
    DialoGPT-FUDGE & 0.6684 & - & 10.26 & 0.60 & 0.0514 & 0.0510 s/tok \\
    DialoGPT-PPLM & 0.6858 & - & 11.30 & 0.94 & 0.0646 & 0.5208 s/tok \\
    DialoGPT-SideControl & \textbf{0.7526} & \textbf{14.34} & 13.46 & 1.96 & 0.0988 & 0.0824 s/tok \\
    \midrule
    BlenderBot-Ori & 0.7455 & 90.89 & 9.38 & 0.54 & 0.0908 & 0.0384 s/tok \\
    BlenderBot-SideControl & \textbf{0.7841} & \textbf{14.34} & 10.10 & 1.20 & 0.0964 & 0.0608 s/tok \\
    \bottomrule
  \end{tabular}
  \caption{\label{tab:kb-result}
   Knowledge document control performances under full training set of ConvAI2, where $\lambda=10^{-5}$ for $\mathcal{L}_{control}$ in DialoGPT-SideControl and BlenderBot-SideControl.
  }
\end{table*}

\begin{table*}[h!]
  \centering
  \small
  \begin{tabular}{lcccccc}
    \toprule
    & \multicolumn{1}{c}{\textbf{Controllability}} & \multicolumn{4}{c}{\textbf{Text Quality}} & \multicolumn{1}{c}{\textbf{Decoding Cost}} \\
     \cmidrule(lr){2-2} \cmidrule(lr){3-6} \cmidrule(lr){7-7}
    \textsc{Method} & \textsc{Accuracy} $\uparrow$ & \textsc{PPL} $\downarrow$ & \textsc{BLEU-1} $\uparrow$  & \textsc{BLEU-2} $\uparrow$  & \textsc{METEOR} $\uparrow$ & \textsc{Time} $\downarrow$  \\
    \midrule
    DialoGPT-Ori & 0.4307 & 55.09 & 7.78 & 0.66 & 0.0333 & 0.0921 s/tok \\
    DialoGPT-FT & 0.4358 & 8.95 & 14.35 & 2.30 & 0.0523 & 0.0786 s/tok \\
    DialoGPT-FUDGE & 0.4701 & - & 14.40 & 1.59 & 0.0411 & 0.0535 s/tok \\
    DialoGPT-PPLM & 0.5994 & - & 14.22 & 1.25 & 0.0506 &  2.4171 s/tok \\
    DialoGPT-SideControl & 0.5376 & 12.79 & 16.37 & 1.90 & 0.0526 & 0.0990 s/tok \\
    \midrule
    BlenderBot-Ori & 0.4605 & 110.05 & 12.21 & 1.10 & 0.0775 & 0.0603 s/tok \\
    BlenderBot-SideControl & \textbf{0.6865} & \textbf{8.16} & 14.49 & 1.36 & 0.0680 & 0.0995 s/tok \\
    \bottomrule
  \end{tabular}
  \caption{\label{tab:da-result}
  Semantic label control performances under full training set of DailyDialog, where $\lambda=10^5$ for $\mathcal{L}_{control}$ in DialoGPT-SideControl and BlenderBot-SideControl.
  }
\end{table*}

\newparagraph{BlenderBot-Ori}: the original BlenderBot \citep{roller2020recipes}, which is a Transformer-based sequence-to-sequence model showing state-of-the-art performance on some challenging open-domain dialogue datasets.

\newparagraph{BlenderBot-SideControl}: our \textsc{SideControl} framework with BlenderBot as the base pre-trained language model. It is used to show the high modularity of our side network.

\subsection{Knowledge Document Control}

In this task, given the previous dialogue context and the external knowledge document for the current speaker, the model will generate one utterance that is relevant both to the context and to the knowledge document.
We provide the detailed experiment setups in \autoref{sec:app-kb-setup}.

\paragraph{Dataset.}
We use the ConvAI2 dataset \citep{dinan2020second} for the knowledge document control task.
\remove[WD]{The ConvAI2 dataset provides conversations between two crowdworkers chattting with each other based on a randomly assigned persona profile (i.e. a document containing 4-5 sentences that describes various personal interests).}
We set the previous 4 utterances as the dialogue context.
Each utterance is linked to its corresponding persona profile.
Since the test set of ConvAI2 has not been made public, we use the original training set to construct our training set, and split the first 80\% original validation set as our validation set and the remaining 20\% original validation set as our testing set.
In total, we have 153,082 training samples, 38,271 validation samples and 11,590 testing samples.

\begin{figure}[t]
    \centering
    \includegraphics[width=0.45\textwidth]{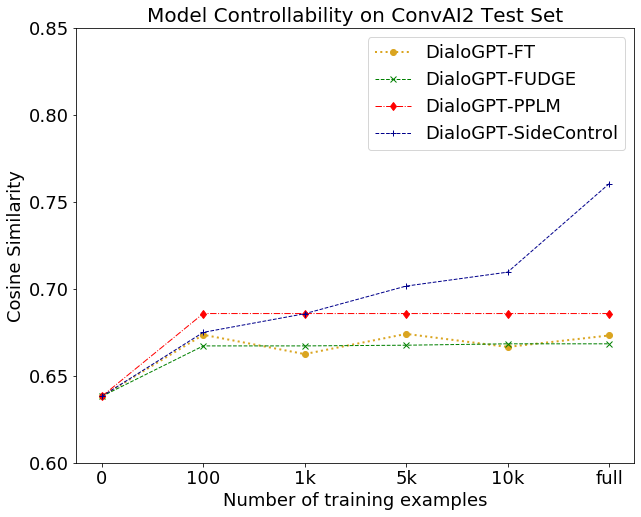}
    \caption{Controllability under different number of training examples in ConvAI2 dataset.}
    \label{fig:kb_sim}
\end{figure}

\paragraph{Performances under Full Data.}
\autoref{tab:kb-result} shows that DialoGPT+{SideControl} outperforms all other baselines in controllability, which validates the effectiveness of the \textsc{SideControl} framework.
For the quality of the generated texts, we find that both FUDGE and PPLM perform worse than the original pre-trained language model, while the \textsc{SideControl} shows improved quality because of the $\mathcal{L}_{cclm}$ during training.
We also notice that direct fine-tuning gives the best performance in BLEU-1 and BLEU-2, but worse controllability compared with the \textsc{SideControl}.
\change[WD]{This is because we explicitly incorporate $\mathcal{L}_{control}$ into the \textsc{SideControl} during training.}{This is because direct fine-tuning only focuses on optimizing the language modelling loss, and does not take the control attributes information into account.}
For the decoding cost, our \textsc{SideControl} is around 6x faster than PPLM during generation, which shows its efficiency during inference.
Finally, we find that the performance improvement in controllability and text quality also hold when we apply the \textsc{SideControl} to BlenderBot, which shows the flexible modularity of the side network.

\paragraph{Performances under Small Data.}
\change[WD]{We also examine the performance of \textsc{SideControl} under different training data size}{With the goal of testing the sample-efficiency of the \textsc{SideControl} framework, we train all baselines under smaller datasets}, where we randomly sample 100, 1000, 5000 and 10000 training samples from the original training set to train the model, and evaluate the model performance using the full testing test. 
\autoref{fig:kb_sim} shows the controllability performance under different training sizes, and we provide detailed text quality performance in \autoref{sec:app-kb-results}.
We find that \textsc{SideControl} only underperforms PPLM in 100 training samples, since PPLM uses non-parametric bag-of-words features as its attribute model while \textsc{SideControl} uses a BiLSTM as its attribute model.
And 1000 training samples are sufficient enough for \textsc{SideControl} to achieve comparable performance with PPLM.
In addition, \textsc{SideControl} constantly achieves performance improvement when increasing the training size.

\paragraph{Ablation Study.}
To verify the effectiveness of the control loss $\mathcal{L}_{control}$, we conduct ablation study by trying out different values of $\lambda$ in \autoref{eq:general-objective}.
We provide partial results in \autoref{tab:ablation-kb-sub} and full results in \autoref{sec:app-ablation-both}.
When $\lambda=0$, the model becomes a vanilla language model and takes no information from the side network, which leads to a low performance in controllability.
When $\lambda\ne0$, the model incorporates control attributes information from the side network, which leads to an improved performance in controllability. However, incorporating side information will lead to a slight increase in model perplexity.

\subsection{Semantic Label Control} 

In this task, given the previous dialogue context and the current dialogue act, the model will generate one utterance that is relevant to the context and also satisfies the current dialogue act. 
We provide the detailed experiment setups in \autoref{sec:app-da-setup}.

\paragraph{Dataset.}
We use the DailyDialog dataset \citep{li-etal-2017-dailydialog} for the semantic label control task.
\remove[WD]{Each utterance is manually annotated with a corresponding dialogue act label, which covers one of the following communication purpose: \textit{inform}, \textit{questions}, \textit{directives} and \textit{commissive}.}
We set the previous 5 utterances as the dialogue context and follow the standard train/validation/test splition of the original dataset to construct our generation dataset.
In total, we obtain 35,781 training samples, 3,388 validation samples and 3,123 testing samples.

\begin{figure}[t]
    \centering
    \includegraphics[width=0.45\textwidth]{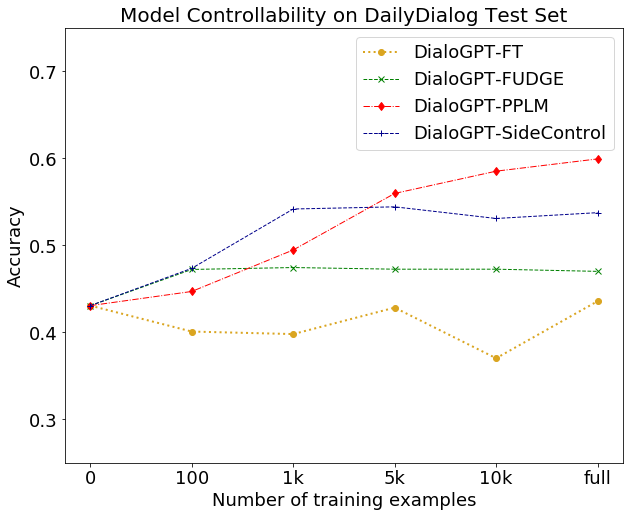}
    \caption{Controllability under different number of training examples in DailyDialog dataset.}
    \label{fig:da_acc}
\end{figure}

\begin{table*}[h!]
\parbox{.45\linewidth}{
\centering
\small
\begin{tabular}{lcc}
\hline
$\lambda$ & \textsc{Similarity} $\uparrow$ & \textsc{PPL} $\downarrow$ \\
\hline
$\lambda=0$ & 0.7273 & \textbf{14.24} \\
$\lambda={10}^{-4}$ & 0.7306 & 14.65 \\
$\lambda={10}^{-5}$ & \textbf{0.7526} & 14.34 \\
\hline
\end{tabular}
\caption{
\label{tab:ablation-kb-sub}
Ablation study of DialoGPT-SideControl on knowledge document control, where the model is trained under the full training set of ConvAI2, and tested under the full testing set of ConvAI2.}
}
\hfill
\parbox{.45\linewidth}{
\centering
\small
\begin{tabular}{lcc}
\hline
$\lambda$ & \textsc{Accuracy} $\uparrow$ & \textsc{PPL} $\downarrow$ \\
\hline
$\lambda=0$ & 0.4950 & \textbf{12.37} \\
$\lambda={10}^3$ & 0.5232 & 12.59 \\
$\lambda={10}^5$ & \textbf{0.5376} & 12.79 \\
\hline
\end{tabular}
\caption{
\label{tab:ablation-da-sub}
Ablation study of DialoGPT-SideControl on semantic label control, where the model is trained under the full training set of DailyDialog, and tested under the full testing set of DailyDialog.}
}
\end{table*}
\begin{table*}[h!]
\parbox{.45\linewidth}{
\centering
\small
\begin{tabular}{lcc}
\hline
\textsc{Method} & \textsc{Fluency} $\uparrow$ & \textsc{Relevancy} $\uparrow$ \\
\hline
DialoGPT-PPLM & 3.832 & 3.188 \\
DialoGPT-FUDGE & 4.016 & 3.348 \\
DialoGPT-SideControl & \textbf{4.108} & \textbf{3.816} \\
\hline
\end{tabular}
\caption{
\label{tab:human-eval1}
Human evaluation of fluency and context relevancy on semantic label control task.}
}
\hfill
\parbox{.45\linewidth}{
\centering
\small
\begin{tabular}{lccc}
\hline
\multicolumn{1}{l}{} & \multicolumn{3}{c}{Wins \%} \\
\cmidrule(lr){2-4}
& PPLM & FUDGE & SideControl \\
\hline
PPLM & - & 55.25\% & \textbf{57.61\%} \\
FUDGE & 44.76\% & - & \textbf{54.46\%} \\
SideControl & 42.39\% & 45.54\% & - \\
\hline
\end{tabular}
\caption{
\label{tab:human-eval2}
Human evaluation of attribute relevancy on semantic label control task.}
}
\end{table*}

\paragraph{Performances under Full Data.}
\autoref{tab:da-result} demonstrates that \textsc{SideControl} has better text quality than FUDGE and PPLM, since we explicitly optimize $\mathcal{L}_{cclm}$ during training.
For the controllability, PPLM achieves the best performance with a sacrifice of inference efficiency, while \textsc{SideControl} can achieve comparable performance in controllability with around 24x faster decoding time.
Finally, the performance improvements in controllability and text quality still hold when we switch the base network from DialoGPT to BlenderBot, which demonstrates that the side network is flexible to be applied to \change[WD]{other pre-trained models}{different types of pre-trained language models}.
And surprisingly, BlenderBot can even provide the state-of-the-art performance in controllability.

\paragraph{Performances under Small Data.}
We also compare across the model performance under different training sizes following the same setup with the knowledge document control task, and provide detailed text quality performance in \autoref{sec:app-da-results}.
\autoref{fig:da_acc} illustrates that \textsc{SideControl} achieves better controllability than PPLM when training size is under 1000.
This is because PPLM uses a data-driven classifier as its attribute model in this task, and its attribute model gets overfitted on the 100 training samples, which results in poor controllability performance.
{Similarly, FUDGE has the same overfitting issue for its attribute discriminator on these small training sets, and gets unsatisfied controllability performance.}
Although \textsc{SideControl} also pre-trains a classifier on the 100 training samples to guide the update of side representation, its final representation is a combination of base and side representation.
We believe incorporating prior knowledge from the base representation helps \textsc{SideControl} alleviates the overfitting issue on small training set.

\paragraph{Ablation Study.}
We also try out different values of $\lambda$ to study the effect of control loss $\mathcal{L}_{control}$, as shown in \autoref{tab:ablation-da-sub}. Full ablation study results are provided in \autoref{sec:app-ablation-both}.
When $\lambda=0$, the model takes no control attributes signals from the side network during training, which results in a low controllability performance.
When $\lambda\ne0$, the controllability performance of the model is improved but with a slight increase in model perplexity. 
Both \autoref{tab:ablation-kb-sub} and \autoref{tab:ablation-da-sub} verify the effectiveness of control loss $\mathcal{L}_{control}$ in improving the controllability of pre-trained language models. 

\paragraph{Human Evaluation.}
To validate the good performance of \textsc{SideControl}, we follow prior works \citep{DBLP:journals/corr/abs-1912-02164,DBLP:journals/corr/abs-1909-03087} and deploy a set of human evaluations to compare the text quality and controllability between several methods. 
For the text quality, we ask human annotators to evaluate the fluency and context relevancy of the generated responses on a scale of 1-5, where a higher score indicates better quality.
For the controllability, we use A/B testing following \citep{DBLP:journals/corr/abs-1909-03087} and compare all model pairs (e.g. PPLM vs. \textsc{SideControl}) \footnote{We show the same dialogue context, current dialog act and two responses generated by model A and model B respectively, and ask human annotators to select the response which is more related to the current dialog act among: model A, model B, both and neither.}.
For all human evaluations, we randomly sample 50 dialogue contexts, and collect the corresponding model generated responses. Human annotators are recruited using Amazon Mechanical Turk and each response has 5 annotations. In total, we collect 2250 human annotations.
\autoref{tab:human-eval1} shows the results of text quality evaluation, and \textsc{SideControl} achieves the best fluency and context relevancy than PPLM and FUDGE.
\autoref{tab:human-eval2} shows the results of controllability evaluation, and \textsc{SideControl} wins over PPLM and FUDGE in 57\% and 54\% respectively.
Both text quality and controllability evaluation show that \textsc{SideControl} can generate more fluent, context-relevant and attribute-relevant responses than PPLM and FUDGE. 

\section{Related Works}
\label{sec:related}




There are three major categories of controllable text generation models:
class-conditional language model~\citep{DBLP:journals/corr/abs-1909-05858, kawano-etal-2019-neural}, plug-and-play language model~\citep{DBLP:journals/corr/abs-1912-02164} and weighted decoding~\citep{ghazvininejad-etal-2017-hafez,baheti-etal-2018-generating,holtzman-etal-2018-learning,yang-klein-2021-fudge}.

\paragraph{Class-Conditional Language Model.}
Class-conditional language models train a conditional generative model from scratch, and guide the generation with explicit control codes provided in the training data.
\citet{DBLP:journals/corr/abs-1909-05858} trains a 1.63 billion-parameter Conditional Transformer Language (CTRL) model by prepending control codes in front of raw texts. 
But training the CTRL \citep{DBLP:journals/corr/abs-1909-05858} requires 140 GB of training data, which may not be affordable for some low-resource languages.
\citet{kawano-etal-2019-neural} builds a controllable neural conversation model by leveraging an adversarial learning framework that alternatively trains between a class-conditional language model and a multi-class discriminator, where the discriminator is used to help the generative model produce responses with appropriate dialogue act.  
But the control code is modelled as discrete variable in this work, which limits the controllability capacity of the dialogue model.

\paragraph{Plug-and-Play Language Model.}
Guiding generation with gradients from additional attribute models is another popular approach.
\citet{DBLP:journals/corr/abs-1912-02164} introduce a plug-and-play language model (PPLM) which combines the pre-trained language model $p(x)$ with attribute models $p(a|x)$ to approximate the contional generative model $p(x|a)$ . At each decoding timestep, all hidden representations of the pre-trained language model are shifted with gradients towards a higher $p(x|a)\propto p(a|x)p(x)$.
The attribute models of PPLM are either in the form of bag-of-words or single layer classifiers, which requires much less training data than learning a conditional generative model.
The following works \citep{goswamy-etal-2020-adapting, DBLP:journals/corr/abs-2104-04039,madotto-etal-2020-plug} further propose more fine-grained attribute models and generation strategies for specific task, such as emotional text generation \citep{goswamy-etal-2020-adapting}, story generation \citep{DBLP:journals/corr/abs-2104-04039} and conversation generation \citep{madotto-etal-2020-plug}.
But since the plug-and-play language models have to compute gradient from attribute model and update hidden representations at each decoding timestep, the generation process is very time-consuming, which leads to high decoding cost.

\paragraph{Weighted Decoding.}
Weighted decoding runs a more expensive beam search where the sampling probability distribution is altered by desired control attributes, such as topic, sentiment, etc.
\citet{ghazvininejad-etal-2017-hafez} design a set of style features on controlling topic, sentiment, and repetitive words, and re-compute the beam score of each token with a combination of the original beam score and the style feature score. 
A recent work \citep{yang-klein-2021-fudge} introduces a Future Discriminator for Generation (FUDGE) that 
trains a binary discriminator for the control attribute prediction and re-scores the probability distribution of the original pre-trained language model with the discriminator prediction via Bayesian factorization.
The major limitation of weighted decoding methods is that, if the pre-trained language model is a high-bias estimator, which assigns low probability for desired attribute words and high probability for commonly observed but unrelated words, re-scoring or re-ranking such a ``high-biased'' distribution cannot guarantee the generation of desired attributes.

The \textsc{SideControl} framework differs the above methods as follows:
(1) the side network only requires access to last hidden states of the base network. Both class-conditional language models \citep{DBLP:journals/corr/abs-1909-05858} and plug-and-play language models \citep{DBLP:journals/corr/abs-1912-02164} require access to every hidden states of the pre-trained language model, which limits its application under certain pre-trained model.
(2) the side network learns a residual on top of pre-trained language models, which is suitable for small datasets. 
Directly fine-tuning \citep{DBLP:journals/corr/abs-1909-08593} large pre-trained language models will cause overfitting issues on some small datasets, and weighted-decoding methods \citep{ghazvininejad-etal-2017-hafez, yang-klein-2021-fudge} only modify the final vocabulary distribution of pre-trained models but do not learn model parameters to better adapt to the target task. 

\section{Conclusions}

In this work, we propose a new method for controlled dialogue generation: adding a small side network to incorporate useful control signals into the pre-trained language models.
We design control attributes loss to teach the side network learning useful control signals.
Empirical experiments show that our method is effective even with $100\sim 1000$ training samples.
Besides, our side network supports diverse forms of attributes control and can be flexibly applied to any pre-trained language models, which extends its possible application to other general controlled text generation tasks. 

\section*{Acknowledgments}
The authors thank the anonymous reviewers for their useful comments and the UVa NLP group for helpful discussions.

\bibliography{anthology,custom}
\bibliographystyle{acl_natbib}

\clearpage

\appendix
\section{Automatic Metrics for Controllability Evaluation}
\label{sec:metrics-control}

In this section we provide implementation details for how we compute classification accuracy and cosine similarity.

\subsection{Dialogue Act Classifier}
We train an independent dialogue act classifier to evaluate whether the current generated response matches its conditioning dialogue act.
The input to the evaluation dialogue act classifier is a single response, and the output is a prediction of one of the 4 dialogues in DailyDialog, i.e. \textit{inform}, \textit{questions}, \textit{directives} and \textit{commissive}.

We construct the training corpus following the standard splition of original DailyDialog dataset, and obtain 87,170 training samples, 8,069 validation samples and 7,740 testing samples.
We leverage the BERT model to provide a sequence of word representations and add a single-layer feed-forward neural network to predict the dialogue act of current sentence.
We use AdamW \citep{loshchilov2019decoupled} with learning rate 0.0001 to train this classifer.
We set the batch size to 16, the total training epoch to 10 and automatically evaluate the model on the validation set very 5000 iterations.
We save the model checkpoint with the lowest validation loss as the optimal model.

\begin{figure}[h]
    \centering
    \includegraphics[width=0.45\textwidth]{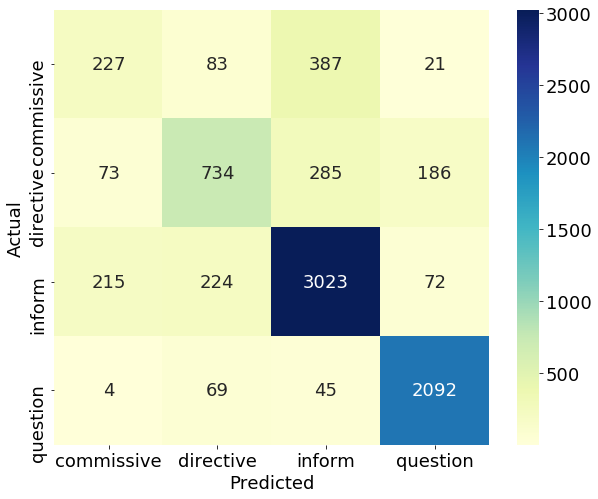}
    \caption{Confusion matrix of the evaluation dialogue act classifier.}
    \label{fig:da_clf}
\end{figure}

This dialogue act classifier achieves 0.79 accuracy on the test set. \autoref{fig:da_clf} shows the confusion matrix of this dialogue act classifier.

\subsection{Computation of Cosine Similarity}
To measure the similarity between the generated response and the conditioning knowledge document, we compute the cosine similarity between the wording embeddings of generated response and external knowledge document.
The word embeddings are GloVe embeddings \citep{pennington-etal-2014-glove} pre-trained on Wikipedia 2014 and Gigaword 5, which are 100-dimension vectors and have 6 billion tokens \footnote{\url{https://nlp.stanford.edu/projects/glove/}}.

We use the NLTK word tokenizer \footnote{\url{https://www.nltk.org/_modules/nltk/tokenize.html}} to tokenize the texts into a set of tokens, and remove stop words based on a pre-defined stop words list in \citep{bao2019plato}.
Finally, we compute the cosine similarity between the two sets of word vectors.

\section{Experiment Setups for Knowledge Document Control}
\label{sec:app-kb-setup}

We conduct all of our experiments on single GeForce RTX 2080Ti GPU server with 11019 MB memory.

\subsection{Direct Fine-tuning}
We directly update all parameters of the pre-trained language model on the ConvAI2 training set without having any side network or control attributes loss.
For the training of the pre-trained language model, we use AdamW \citep{loshchilov2019decoupled} with learning rate 0.0001.
We set the batch size to 2, the total training epoch to 10, and automatically evaluate the model on the validation set every 1000 iterations.
We save the model checkpoint which achieves lowest validation loss as the final optimal model.
For generation, we follow the setup of FUDGE, which use top-$k$ sampling with $k=10$.

\subsection{PPLM}
For the implementation of the attribute model, we use the bag-of-words attribute model proposed in the original paper \citep{DBLP:journals/corr/abs-1912-02164} to encode external knowledge document.
We run the model on the ConvAI2 dataset using the code provided by the original paper: \url{https://github.com/uber-research/PPLM}.
We set the maximum generation length to 50, the number of gradient update steps to 3, the step size to 0.03, the window length to 5, the number of generated sentences to 1, $\gamma_{gm}=0.99$, $\lambda_{KL}=0.01$.

\subsection{FUDGE}
For the implementation of the attribute model, we use the bag-of-words attribute model proposed in the original paper \citep{yang-klein-2021-fudge} to encode external knowledge document.
We run the model on the ConvAI2 dataset using the code provided by the original paper: \url{https://github.com/yangkevin2/naacl-2021-fudge-controlled-generation}.
We set the maximum generation length to 80, the weight on conditioning model to 4.0, consider top 200 outputs from DialoGPT at each decoding timestep before conditioning, and sample from top 10 outputs from DialoGPT at each decoding timestep.

\subsection{SideControl}
For the implementation of the side network, we use a single-layer bi-LSTM which shares the same hidden dimension with the final hidden states of the base network. 
We tokenize the knowledge document using the same tokenizer with the base network, and share the same word embedding with the base network as well.
For the training of the side network, we use AdamW \citep{loshchilov2019decoupled} with learning rate 0.0001.
We set the batch size to 4, the total training epoch to 10, and automatically evaluate the model on the  validation set every 100 iterations.
For the hyper-parameter $\lambda$ of the coverage loss in \autoref{eq:kb-control-loss}, we use grid search on the validation set to obtain the optimal number.
We search from the set $\lambda = \{10^{-6}, 10^{-5}, 10^{-4}, 10^{-3}, 0.01, 0.1\}$ and find $\lambda = 10^{-5}$ yields best performance.
For generation, we follow the setup of FUDGE,  which use top-$k$ sampling with $k=10$.

\section{Experiment Setups for Semantic Label Control}
\label{sec:app-da-setup}

We conduct all of our experiments on single GeForce RTX 2080Ti GPU server with 11019 MB memory.

\subsection{Direct Fine-tuning}
We directly update all parameters of the pre-trained language model on the DailyDialog training set without having any side network or control attributes loss.
For the training of the pre-trained language model, we use AdamW \citep{loshchilov2019decoupled} with learning rate 0.0001.
We set the batch size to 2, the total training epoch to 10, and automatically evaluate the model on the validation set every 1000 iterations.
We save the model checkpoint which achieves lowest validation loss as the final optimal model.
For generation, we follow the setup of FUDGE, which use top-$k$ sampling with $k=10$.

\subsection{PPLM}
For the implementation of the attribute model, we follow the generic discriminator implementation in the original paper \citep{DBLP:journals/corr/abs-1912-02164}.
We run the model on the DailyDialog dataset using the code provided by the original paper.
We train a dialogue act classifier which takes single response as input and produces a prediction on one of the four dialogue acts.
For the training of the classifier, we use Adam \citep{kingma2017adam} with learning rate 0.0001.
We set the batch size to 64, the total training epoch to 10.
For the generation of PPLM, we set the maximum generation length to 50, the number of gradient update steps to 10, the step size to 0.2, the number of generated sentences to 1, $\gamma_{gm}=0.95$, $\lambda_{KL}=0.01$.

\subsection{FUDGE}
For the implementation of the attribute model, we follow the attribute discriminator implementation in the original paper \citep{yang-klein-2021-fudge}.
We run the model on the DailyDialog dataset using the code provided by the original paper.
We train a dialogue act discriminator which takes the dialogue context and the current response as input and produces a prediction on one of the four dialogue acts.
For the training of the discriminator, we use Adam \citep{kingma2017adam} with learning rate $2 \times 10^{-5}$.
We set the batch size to 16, the total training epoch to 10.
For the generation of FUDGE, we set the maximum generation length to 60, the weight on conditioning model to 1.0, consider top 200 outputs from DialoGPT at each decoding timestep before conditioning, and sample from top 10 outputs from DialoGPT at each decoding timestep.

\subsection{SideControl}
For the implementation of the side network, we use a single-layer feed-forward neural network which shares the same hidden dimension with the final hidden states of the base network. 
Besides, we pre-trained a dialogue act classifier to compute the control loss in \autoref{eq:da-control-loss}.
We emphasize that this dialogue act classifier is different from the evaluation classifier. 
It models the sentence representation from the base network, i.e. DialoGPT, and adds a single-layer feed-forward neural network to predict the dialogue act of current response.
We train this classifier using AdamW \citep{loshchilov2019decoupled} with learning rate 0.0001 for 10 epochs. 
Then, we fix this classifier and begin to train the side network using AdamW \citep{loshchilov2019decoupled} with learning rate 0.0001 for another 10 epochs. 
We evaluate the model on the validation set every 1000 iterations, and save the model checkpoint which has the lowest validation loss.
For the hyper-parameter $\lambda$ of the control loss in \autoref{eq:da-control-loss}, we use grid search on the validation set to obtain the optimal number.
We search from the set $\lambda = \{1, 10, 100, 10^3, 10^4, 10^5, 10^6\}$ and find $\lambda = 10^{5}$ yields best performance on the full training set.
For generation, we follow the setup of FUDGE,  which use top-$k$ sampling with $k=10$.

\section{Full performances of Ablation Study}
\label{sec:app-ablation-both}
We provide performance details for ablation study in knowledge document control and semantic label control.
The full performances of ablation study in knowledge document control is shown in \autoref{tab:kb-ablation}.
The full performances of ablation study in semantic label control is shown in \autoref{tab:da-ablation}.

\begin{table*}[h!]
  \centering
  \small
  \begin{tabular}{lccccc}
    \toprule
    & \multicolumn{1}{c}{\textbf{Controllability}} & \multicolumn{4}{c}{\textbf{Text Quality}}  \\
     \cmidrule(lr){2-2} \cmidrule(lr){3-6} 
     & \textsc{Similarity} $\uparrow$ & \textsc{Perplexity} $\downarrow$ & \textsc{BLEU-1} $\uparrow$  & \textsc{BLEU-2} $\uparrow$  & \textsc{METEOR} $\uparrow$  \\
    \midrule
    $\lambda=0$ & 0.7273 & \textbf{14.24} & 15.72 & 2.16 & 0.0858 \\
    $\lambda={10}^{-6}$ & 0.7284 & 14.30 & 16.08 & 2.29 & 0.0800 \\
    $\lambda={10}^{-5}$ & \textbf{0.7526} & 14.34 & 13.46 & 1.96 & 0.0988 \\
    $\lambda={10}^{-4}$ & 0.7306 & 14.65 & 15.87 & 2.32 & 0.0846 \\
    $\lambda={10}^{-3}$ & 0.7259 & 15.65 & 15.72 & 2.09 & 0.0802 \\
    $\lambda={10}^{-2}$ & 0.7217 & 30.29 & 15.30 & 2.05 & 0.0803 \\
    $\lambda={10}^{-1}$ & 0.7137 & 22481.68 & 15.50 & 2.01 & 0.0774 \\
    \bottomrule
  \end{tabular}
  \caption{\label{tab:kb-ablation}
  Knowledge document control performances of DialoGPT-SideControl with different $\lambda$.
  }
\end{table*}

\begin{table*}[h!]
  \centering
  \small
  \begin{tabular}{lccccc}
    \toprule
    & \multicolumn{1}{c}{\textbf{Controllability}} & \multicolumn{4}{c}{\textbf{Text Quality}}  \\
     \cmidrule(lr){2-2} \cmidrule(lr){3-6} 
    & \textsc{Accuracy} $\uparrow$ & \textsc{Perplexity} $\downarrow$ & \textsc{BLEU-1} $\uparrow$  & \textsc{BLEU-2} $\uparrow$  & \textsc{METEOR} $\uparrow$  \\
    \midrule
    $\lambda=0$ & 0.4950 & \textbf{12.37} & 16.19 & 1.95 & 0.0534 \\
    $\lambda={10}^{1}$ & 0.5229 & 12.48 & 15.06 & 1.76 & 0.0525 \\
    $\lambda={10}^{2}$ & 0.5366 &  12.51 & 15.59 & 1.76 & 0.0517 \\
    $\lambda={10}^{3}$ & 0.5232 & 12.59 & 15.59 & 1.75 & 0.0512 \\
    $\lambda={10}^{5}$ & \textbf{0.5376} & 12.79 & 16.37 & 1.90 & 0.0526 \\
    $\lambda={10}^{6}$ & 0.5357 & 13.10 & 15.29 & 1.67 & 0.0485 \\
    \bottomrule
  \end{tabular}
  \caption{\label{tab:da-ablation}
  Semantic label control performances of DialoGPT-SideControl with different $\lambda$.
  }
\end{table*}

\section{Full performances of Knowledge Document Control under Different Number of Training Samples}
\label{sec:app-kb-results}
For all experiments across different number of training samples, we take the hyper-parameter $\lambda=10^{-5}$ for $\mathcal{L}_{control}$.
Full performance for all models are demonstrated in \autoref{tab:kb-result-100}, \autoref{tab:kb-result-1k}, \autoref{tab:kb-result-5k} and \autoref{tab:kb-result-10k}.
We also provide some generated samples from the test set for reference, demonstrated in \autoref{tab:kb-case-100}, \autoref{tab:kb-case-1k}, \autoref{tab:kb-case-5k}, \autoref{tab:kb-case-10k}, \autoref{tab:kb-case-full}.

\section{Full performances of Semantic Label Control under Different Number of Training Samples}
\label{sec:app-da-results}
For the semantic label control task, we find the optimal hyper-parameter $\lambda$ for $\mathcal{L}_{control}$ differs across different number of training samples.
Full performance for all models are demonstrated in \autoref{tab:da-result-100}, \autoref{tab:da-result-1k}, \autoref{tab:da-result-5k} and \autoref{tab:da-result-10k}.
We also provide some generated samples from the test set for reference, demonstrated in \autoref{tab:da-case-100}, \autoref{tab:da-case-1k}, \autoref{tab:da-case-5k}, \autoref{tab:da-case-10k}, \autoref{tab:da-case-full}.

\begin{table*}[h!]
  \centering
  \small
  \begin{tabular}{lccccc}
    \toprule
    & \multicolumn{1}{c}{\textbf{Controllability}} & \multicolumn{4}{c}{\textbf{Text Quality}}  \\
     \cmidrule(lr){2-2} \cmidrule(lr){3-6} 
    \textsc{Method} & \textsc{Similarity} $\uparrow$ & \textsc{Perplexity} $\downarrow$ & \textsc{BLEU-1} $\uparrow$  & \textsc{BLEU-2} $\uparrow$  & \textsc{METEOR} $\uparrow$  \\
    \midrule
    DialoGPT-FT & 0.6736 & 180.65 & 13.04 & 1.32 & 0.0503  \\
    DialoGPT-FUDGE & 0.6672 & - & 10.33 & 0.56 & 0.0530  \\
    DialoGPT-PPLM & 0.6858 & - & 11.30 & 0.94 & 0.0646  \\
    DialoGPT-SideControl & 0.6750 & 24.64 & 12.00 & 1.31 & 0.0553  \\
    \bottomrule
  \end{tabular}
  \caption{\label{tab:kb-result-100}
  Knowledge document control performances under 100 training samples sampled from ConvAI2, where $\lambda=10^{-5}$ for $\mathcal{L}_{control}$ in DialoGPT-SideControl.
  }
\end{table*}

\begin{table*}[h!]
  \centering
  \small
  \begin{tabular}{p{0.55\textwidth}|p{0.35\textwidth}}
    \toprule
    \textbf{Dialogue Context}: & \textbf{Persona Profile}: \\
    ``hi how are you tonight ? i am relaxing in a bubble bath tonight .'' &  ``my favorite season is winter.'' \\
    ``hi how are you doing that sounds great .'' & ``i won a gold medal in the 1992 olympics.'' \\
    ``it is , how old are you ? i'm 31 , feeling like i'm getting old lol.'' & ``i love cats and have two cats.'' \\
    ``not old you are still young try 50 i'm getting old .'' & ``i've a hat collection of over 1000 hats.'' \\
    ``lol my bones suggest other wise . do you have any animals ?'' & ``i have a weakness for vanilla ice cream.'' \\
    \midrule
    \multicolumn{2}{l}{\textbf{Ground Truth:} ''yes i like and love all animals i have two cats.''} \\
    \multicolumn{2}{l}{\textbf{DialoGPT-Ori:} ''my dogs are my favorite, but i'm pretty fond of all dogs so i don't have many.''} \\
    \multicolumn{2}{l}{\textbf{DialoGPT-FT:} ''i love my three dogs. ''} \\
    \multicolumn{2}{l}{\textbf{DialoGPT-FUDGE:} ''My bones say I'm young too, so I think we're good there for now.''} \\
    \multicolumn{2}{l}{\textbf{DialoGPT-PPLM:} ''I think i'm only 35, but i've been in a pretty good relationship so i've probably gotten better.''} \\
    \multicolumn{2}{l}{\textbf{DialoGPT-SideControl:} ''!''} \\
    \bottomrule
  \end{tabular}
  \caption{\label{tab:kb-case-100}
  Knowledge document control generation results under 100 training samples sampled from ConvAI2.
  }
\end{table*}

\begin{table*}[h!]
  \centering
  \small
  \begin{tabular}{lccccc}
    \toprule
    & \multicolumn{1}{c}{\textbf{Controllability}} & \multicolumn{4}{c}{\textbf{Text Quality}}  \\
     \cmidrule(lr){2-2} \cmidrule(lr){3-6} 
    \textsc{Method} & \textsc{Similarity} $\uparrow$ & \textsc{Perplexity} $\downarrow$ & \textsc{BLEU-1} $\uparrow$  & \textsc{BLEU-2} $\uparrow$  & \textsc{METEOR} $\uparrow$  \\
    \midrule
    DialoGPT-FT & 0.6625 & 18.64 & 15.67 & 1.69 & 0.0628  \\
    DialoGPT-FUDGE & 0.6672 & - & 10.33 & 0.56 & 0.0530  \\
    DialoGPT-PPLM & 0.6858 & - & 11.30 & 0.94 & 0.0646  \\
    DialoGPT-SideControl & 0.6857 & 19.32 & 15.88 & 1.98 & 0.0748  \\
    \bottomrule
  \end{tabular}
  \caption{\label{tab:kb-result-1k}
  Knowledge document control performances under 1000 training samples sampled from ConvAI2, where $\lambda=10^{-5}$ for $\mathcal{L}_{control}$ in DialoGPT-SideControl.
  }
\end{table*}

\begin{table*}[h!]
  \centering
  \small
  \begin{tabular}{p{0.55\textwidth}|p{0.35\textwidth}}
    \toprule
    \textbf{Dialogue Context}: & \textbf{Persona Profile}: \\
    ``hi how are you tonight ? i am relaxing in a bubble bath tonight .'' &  ``my favorite season is winter.'' \\
    ``hi how are you doing that sounds great .'' & ``i won a gold medal in the 1992 olympics.'' \\
    ``it is , how old are you ? i'm 31 , feeling like i'm getting old lol.'' & ``i love cats and have two cats.'' \\
    ``not old you are still young try 50 i'm getting old .'' & ``i've a hat collection of over 1000 hats.'' \\
    ``lol my bones suggest other wise . do you have any animals ?'' & ``i have a weakness for vanilla ice cream.'' \\
    \midrule
    \multicolumn{2}{l}{\textbf{Ground Truth:} ''yes i like and love all animals i have two cats.''} \\
    \multicolumn{2}{l}{\textbf{DialoGPT-Ori:} ''my dogs are my favorite, but i'm pretty fond of all dogs so i don't have many.''} \\
    \multicolumn{2}{l}{\textbf{DialoGPT-FT:} ''cats and dogs. you?''} \\
    \multicolumn{2}{l}{\textbf{DialoGPT-FUDGE:} ''My bones say I'm young too, so I think we're good there for now.''} \\
    \multicolumn{2}{l}{\textbf{DialoGPT-PPLM:} ''I think i'm only 35, but i've been in a pretty good relationship so i've probably gotten better.''} \\
    \multicolumn{2}{l}{\textbf{DialoGPT-SideControl:} ''i don't, they don't get to play.''} \\
    \bottomrule
  \end{tabular}
  \caption{\label{tab:kb-case-1k}
  Knowledge document control generation results under 1000 training samples sampled from ConvAI2.
  }
\end{table*}

\begin{table*}[h!]
  \centering
  \small
  \begin{tabular}{lccccc}
    \toprule
    & \multicolumn{1}{c}{\textbf{Controllability}} & \multicolumn{4}{c}{\textbf{Text Quality}}  \\
     \cmidrule(lr){2-2} \cmidrule(lr){3-6} 
    \textsc{Method} & \textsc{Similarity} $\uparrow$ & \textsc{Perplexity} $\downarrow$ & \textsc{BLEU-1} $\uparrow$  & \textsc{BLEU-2} $\uparrow$  & \textsc{METEOR} $\uparrow$  \\
    \midrule
    DialoGPT-FT & 0.6741 & 24.46 & 15.65 & 1.17 & 0.0634  \\
    DialoGPT-FUDGE & 0.6676 & - & 10.20 & 0.54 & 0.0528  \\
    DialoGPT-PPLM & 0.6858 & - & 11.30 & 0.94 & 0.0646  \\
    DialoGPT-SideControl & 0.7016 & 17.30 & 15.76 & 1.99 & 0.0764  \\
    \bottomrule
  \end{tabular}
  \caption{\label{tab:kb-result-5k}
  Knowledge document control performances under 5000 training samples sampled from ConvAI2, where $\lambda=10^{-5}$ for $\mathcal{L}_{control}$ in DialoGPT-SideControl.
  }
\end{table*}

\begin{table*}[h!]
  \centering
  \small
  \begin{tabular}{p{0.55\textwidth}|p{0.35\textwidth}}
    \toprule
    \textbf{Dialogue Context}: & \textbf{Persona Profile}: \\
    ``hi how are you tonight ? i am relaxing in a bubble bath tonight .'' &  ``my favorite season is winter.'' \\
    ``hi how are you doing that sounds great .'' & ``i won a gold medal in the 1992 olympics.'' \\
    ``it is , how old are you ? i'm 31 , feeling like i'm getting old lol.'' & ``i love cats and have two cats.'' \\
    ``not old you are still young try 50 i'm getting old .'' & ``i've a hat collection of over 1000 hats.'' \\
    ``lol my bones suggest other wise . do you have any animals ?'' & ``i have a weakness for vanilla ice cream.'' \\
    \midrule
    \multicolumn{2}{l}{\textbf{Ground Truth:} ''yes i like and love all animals i have two cats.''} \\
    \multicolumn{2}{l}{\textbf{DialoGPT-Ori:} ''my dogs are my favorite, but i'm pretty fond of all dogs so i don't have many.''} \\
    \multicolumn{2}{l}{\textbf{DialoGPT-FT:} ''!!! i've 3 cats but they are all scared of me haha.''} \\
    \multicolumn{2}{l}{\textbf{DialoGPT-FUDGE:} ''My bones say I'm young too, so I think we're good there for now.''} \\
    \multicolumn{2}{l}{\textbf{DialoGPT-PPLM:} ''I think i'm only 35, but i've been in a pretty good relationship so i've probably gotten better.''} \\
    \multicolumn{2}{l}{\textbf{DialoGPT-SideControl:} ''i do. i love them. what do you do for fun? goldfish?''} \\
    \bottomrule
  \end{tabular}
  \caption{\label{tab:kb-case-5k}
  Knowledge document control generation results under 5000 training samples sampled from ConvAI2.
  }
\end{table*}

\begin{table*}[h!]
  \centering
  \small
  \begin{tabular}{lccccc}
    \toprule
    & \multicolumn{1}{c}{\textbf{Controllability}} & \multicolumn{4}{c}{\textbf{Text Quality}}  \\
     \cmidrule(lr){2-2} \cmidrule(lr){3-6} 
    \textsc{Method} & \textsc{Similarity} $\uparrow$ & \textsc{Perplexity} $\downarrow$ & \textsc{BLEU-1} $\uparrow$  & \textsc{BLEU-2} $\uparrow$  & \textsc{METEOR} $\uparrow$  \\
    \midrule
    DialoGPT-FT & 0.6666 & 17.78 & 14.86 & 1.56 & 0.0657  \\
    DialoGPT-FUDGE & 0.6684 & - & 10.21 & 0.58 & 0.0529  \\
    DialoGPT-PPLM & 0.6858 & - & 11.30 & 0.94 & 0.0646  \\
    DialoGPT-SideControl & 0.7096 & 16.98 & 15.49 & 1.91 & 0.0774  \\
    \bottomrule
  \end{tabular}
  \caption{\label{tab:kb-result-10k}
  Knowledge document control performances under 10000 training samples sampled from ConvAI2, where $\lambda=10^{-5}$ for $\mathcal{L}_{control}$ in DialoGPT-SideControl.
  }
\end{table*}

\begin{table*}[h!]
  \centering
  \small
  \begin{tabular}{p{0.55\textwidth}|p{0.35\textwidth}}
    \toprule
    \textbf{Dialogue Context}: & \textbf{Persona Profile}: \\
    ``hi how are you tonight ? i am relaxing in a bubble bath tonight .'' &  ``my favorite season is winter.'' \\
    ``hi how are you doing that sounds great .'' & ``i won a gold medal in the 1992 olympics.'' \\
    ``it is , how old are you ? i'm 31 , feeling like i'm getting old lol.'' & ``i love cats and have two cats.'' \\
    ``not old you are still young try 50 i'm getting old .'' & ``i've a hat collection of over 1000 hats.'' \\
    ``lol my bones suggest other wise . do you have any animals ?'' & ``i have a weakness for vanilla ice cream.'' \\
    \midrule
    \multicolumn{2}{l}{\textbf{Ground Truth:} ''yes i like and love all animals i have two cats.''} \\
    \multicolumn{2}{l}{\textbf{DialoGPT-Ori:} ''my dogs are my favorite, but i'm pretty fond of all dogs so i don't have many.''} \\
    \multicolumn{2}{l}{\textbf{DialoGPT-FT:} ''i have a dog! what do you do for work?''} \\
    \multicolumn{2}{l}{\textbf{DialoGPT-FUDGE:} ''My bones say I'm young too, so I think we're good there for now.''} \\
    \multicolumn{2}{l}{\textbf{DialoGPT-PPLM:} ''I think i'm only 35, but i've been in a pretty good relationship so i've probably gotten better.''} \\
    \multicolumn{2}{l}{\textbf{DialoGPT-SideControl:} ''i've a pet cat named monkey.''} \\
    \bottomrule
  \end{tabular}
  \caption{\label{tab:kb-case-10k}
  Knowledge document control generation results under 10000 training samples sampled from ConvAI2.
  }
\end{table*}

\begin{table*}[h!]
  \centering
  \small
  \begin{tabular}{p{0.55\textwidth}|p{0.35\textwidth}}
    \toprule
    \textbf{Dialogue Context}: & \textbf{Persona Profile}: \\
    ``hi how are you tonight ? i am relaxing in a bubble bath tonight .'' &  ``my favorite season is winter.'' \\
    ``hi how are you doing that sounds great .'' & ``i won a gold medal in the 1992 olympics.'' \\
    ``it is , how old are you ? i'm 31 , feeling like i'm getting old lol.'' & ``i love cats and have two cats.'' \\
    ``not old you are still young try 50 i'm getting old .'' & ``i've a hat collection of over 1000 hats.'' \\
    ``lol my bones suggest other wise . do you have any animals ?'' & ``i have a weakness for vanilla ice cream.'' \\
    \midrule
    \multicolumn{2}{l}{\textbf{Ground Truth:} ''yes i like and love all animals i have two cats.''} \\
    \multicolumn{2}{l}{\textbf{DialoGPT-Ori:} ''my dogs are my favorite, but i'm pretty fond of all dogs so i don't have many.''} \\
    \multicolumn{2}{l}{\textbf{DialoGPT-FT:} ''my dogs and cats are old as well.''} \\
    \multicolumn{2}{l}{\textbf{DialoGPT-FUDGE:} ''.''} \\
    \multicolumn{2}{l}{\textbf{DialoGPT-PPLM:} ''I think i'm only 35, but i've been in a pretty good relationship so i've probably gotten better.''} \\
    \multicolumn{2}{p{0.95\textwidth}}{\textbf{DialoGPT-SideControl:} ''! i do not, i love animals. i don't know how to have pets, i'm too busy. i have two cats. they are my best friend.''} \\
    \bottomrule
  \end{tabular}
  \caption{\label{tab:kb-case-full}
  Knowledge document control generation results under full training samples from ConvAI2.
  }
\end{table*}

\begin{table*}[h!]
  \centering
  \small
  \begin{tabular}{lccccc}
    \toprule
    & \multicolumn{1}{c}{\textbf{Controllability}} & \multicolumn{4}{c}{\textbf{Text Quality}}  \\
     \cmidrule(lr){2-2} \cmidrule(lr){3-6} 
    \textsc{Method} & \textsc{Accuracy} $\uparrow$ & \textsc{Perplexity} $\downarrow$ & \textsc{BLEU-1} $\uparrow$  & \textsc{BLEU-2} $\uparrow$  & \textsc{METEOR} $\uparrow$  \\
    \midrule
    DialoGPT-FT & 0.4009 & 70.74 & 9.31 & 0.76 & 0.0364  \\
    DialoGPT-FUDGE & 0.4723 & - & 14.59 & 1.59 & 0.0424  \\
    DialoGPT-PPLM & 0.4470 & - & 11.11 & 0.57 & 0.0382  \\
    DialoGPT-SideControl & 0.4736 & 30.33 & 11.17 & 0.77 & 0.0379  \\
    \bottomrule
  \end{tabular}
  \caption{\label{tab:da-result-100}
  Semantic label control performances under 100 training samples sampled from DailyDialog, where $\lambda=1$ for $\mathcal{L}_{control}$ in DialoGPT-SideControl.
  }
\end{table*}

\begin{table*}[h!]
  \centering
  \small
  \begin{tabular}{p{0.65\textwidth}|p{0.15\textwidth}}
    \toprule
    \textbf{Dialogue Context}: & \textbf{Dialogue Act}: \\
    ``Well , have you ever heard of Oulong tea and compressed tea ?'' &  ``directive'' \\
    ``Oh , yeah . Oulong tea is good for one's health . isn't it ?'' &  \\
    ``You surely know a lot about Chinese tea.'' &  \\
    ``Sure , I like drinking tea at teahouses.'' &  \\
    ``Oh , so do I.'' & \\
    \midrule
    \multicolumn{2}{l}{\textbf{Ground Truth:} ''Why don't we go for one now ?''} \\
    \multicolumn{2}{l}{\textbf{DialoGPT-Ori:} ''I'm sure.''} \\
    \multicolumn{2}{l}{\textbf{DialoGPT-FT:} ''yes. I have a large appetite. I like eating. I like cooking. I like shopping. I like to go to the markets. ''} \\
    \multicolumn{2}{l}{\textbf{DialoGPT-FUDGE:} ''I'm not sure if you're being serious or not, but I'm pretty sure that's not what he's saying.''} \\
    \multicolumn{2}{l}{\textbf{DialoGPT-PPLM:} ''I know that tea from somewhere, so it must be Chinese.''} \\
    \multicolumn{2}{l}{\textbf{DialoGPT-SideControl:} ''You know your way, I can't be sure.''} \\
    \bottomrule
  \end{tabular}
  \caption{\label{tab:da-case-100}
  Semantic label control generation results under 100 training samples sampled from DailyDialog.
  }
\end{table*}

\begin{table*}[h!]
  \centering
  \small
  \begin{tabular}{lccccc}
    \toprule
    & \multicolumn{1}{c}{\textbf{Controllability}} & \multicolumn{4}{c}{\textbf{Text Quality}}  \\
     \cmidrule(lr){2-2} \cmidrule(lr){3-6} 
    \textsc{Method} & \textsc{Accuracy} $\uparrow$ & \textsc{Perplexity} $\downarrow$ & \textsc{BLEU-1} $\uparrow$  & \textsc{BLEU-2} $\uparrow$  & \textsc{METEOR} $\uparrow$  \\
    \midrule
    DialoGPT-FT & 0.3980 & 17.74 & 11.47 & 0.99 & 0.0428  \\
    DialoGPT-FUDGE & 0.4745 & - & 14.58 & 1.61 & 0.0424  \\
    DialoGPT-PPLM & 0.4947 & - & 11.35 & 0.75 & 0.0401  \\
    DialoGPT-SideControl & 0.5418 & 18.74 & 14.38 & 1.38 & 0.0472  \\
    \bottomrule
  \end{tabular}
  \caption{\label{tab:da-result-1k}
  Semantic label control performances under 1000 training samples sampled from DailyDialog, where $\lambda=10^3$ for $\mathcal{L}_{control}$ in DialoGPT-SideControl.
  }
\end{table*}

\begin{table*}[h!]
  \centering
  \small
  \begin{tabular}{p{0.65\textwidth}|p{0.15\textwidth}}
    \toprule
    \textbf{Dialogue Context}: & \textbf{Dialogue Act}: \\
    ``Well , have you ever heard of Oulong tea and compressed tea ?'' &  ``directive'' \\
    ``Oh , yeah . Oulong tea is good for one's health . isn't it ?'' &  \\
    ``You surely know a lot about Chinese tea.'' &  \\
    ``Sure , I like drinking tea at teahouses.'' &  \\
    ``Oh , so do I.'' & \\
    \midrule
    \multicolumn{2}{l}{\textbf{Ground Truth:} ''Why don't we go for one now ?''} \\
    \multicolumn{2}{l}{\textbf{DialoGPT-Ori:} ''I'm sure.''} \\
    \multicolumn{2}{l}{\textbf{DialoGPT-FT:} ''But do you like Chinese tea better than American tea?''} \\
    \multicolumn{2}{l}{\textbf{DialoGPT-FUDGE:} ''I'm not sure if you're being serious or not, but I'm pretty sure that's not what he's saying.''} \\
    \multicolumn{2}{l}{\textbf{DialoGPT-PPLM:} ''I think it's the Chinese version that's for me''} \\
    \multicolumn{2}{l}{\textbf{DialoGPT-SideControl:} ''You are the second person to make my point!''} \\
    \bottomrule
  \end{tabular}
  \caption{\label{tab:da-case-1k}
  Semantic label control generation results under 1000 training samples sampled from DailyDialog.
  }
\end{table*}

\begin{table*}[h!]
  \centering
  \small
  \begin{tabular}{lccccc}
    \toprule
    & \multicolumn{1}{c}{\textbf{Controllability}} & \multicolumn{4}{c}{\textbf{Text Quality}}  \\
     \cmidrule(lr){2-2} \cmidrule(lr){3-6} 
    \textsc{Method} & \textsc{Accuracy} $\uparrow$ & \textsc{Perplexity} $\downarrow$ & \textsc{BLEU-1} $\uparrow$  & \textsc{BLEU-2} $\uparrow$  & \textsc{METEOR} $\uparrow$  \\
    \midrule
    DialoGPT-FT & 0.4284 & 15.41 & 14.51 & 1.4 9& 0.0517  \\
    DialoGPT-FUDGE & 0.4726 & - & 14.63 & 1.62 & 0.0420  \\
    DialoGPT-PPLM & 0.5597 & - & 11.03 & 0.62 & 0.0360  \\
    DialoGPT-SideControl & 0.5443 & 15.38 & 14.17 & 1.55 & 0.0448  \\
    \bottomrule
  \end{tabular}
  \caption{\label{tab:da-result-5k}
  Semantic label control performances under 5000 training samples sampled from DailyDialog, where $\lambda=10^4$ for $\mathcal{L}_{control}$ in DialoGPT-SideControl.
  }
\end{table*}

\begin{table*}[h!]
  \centering
  \small
  \begin{tabular}{p{0.65\textwidth}|p{0.15\textwidth}}
    \toprule
    \textbf{Dialogue Context}: & \textbf{Dialogue Act}: \\
    ``Well , have you ever heard of Oulong tea and compressed tea ?'' &  ``directive'' \\
    ``Oh , yeah . Oulong tea is good for one's health . isn't it ?'' &  \\
    ``You surely know a lot about Chinese tea.'' &  \\
    ``Sure , I like drinking tea at teahouses.'' &  \\
    ``Oh , so do I.'' & \\
    \midrule
    \multicolumn{2}{l}{\textbf{Ground Truth:} ''Why don't we go for one now ?''} \\
    \multicolumn{2}{l}{\textbf{DialoGPT-Ori:} ''I'm sure.''} \\
    \multicolumn{2}{l}{\textbf{DialoGPT-FT:} ''you like the Chinese traditional silk?''} \\
    \multicolumn{2}{l}{\textbf{DialoGPT-FUDGE:} ''I'm not sure if you're being serious or not, but I'm pretty sure that's not what he's saying.''} \\
    \multicolumn{2}{l}{\textbf{DialoGPT-PPLM:} ''It's not tea if you mix tea and tea together''} \\
    \multicolumn{2}{l}{\textbf{DialoGPT-SideControl:} ''That's right. We should go to a tea party.''} \\
    \bottomrule
  \end{tabular}
  \caption{\label{tab:da-case-5k}
  Semantic label control generation results under 5000 training samples sampled from DailyDialog.
  }
\end{table*}

\begin{table*}[h!]
  \centering
  \small
  \begin{tabular}{lccccc}
    \toprule
    & \multicolumn{1}{c}{\textbf{Controllability}} & \multicolumn{4}{c}{\textbf{Text Quality}}  \\
     \cmidrule(lr){2-2} \cmidrule(lr){3-6} 
    \textsc{Method} & \textsc{Accuracy} $\uparrow$ & \textsc{Perplexity} $\downarrow$ & \textsc{BLEU-1} $\uparrow$  & \textsc{BLEU-2} $\uparrow$  & \textsc{METEOR} $\uparrow$  \\
    \midrule
    DialoGPT-FT & 0.3702 & 14.59 & 8.18 & 0.98 & 0.0480  \\
    DialoGPT-FUDGE & 0.4726 & - & 14.52 & 1.59 & 0.0421  \\
    DialoGPT-PPLM & 0.5853 & - & 11.27 & 0.61 & 0.0385  \\
    DialoGPT-SideControl & 0.5309 & 14.36 & 15.33 & 1.61 & 0.0478  \\
    \bottomrule
  \end{tabular}
  \caption{\label{tab:da-result-10k}
  Semantic label control performances under 10000 training samples sampled from DailyDialog, where $\lambda=10^4$ for $\mathcal{L}_{control}$ in DialoGPT-SideControl.
  }
\end{table*}

\begin{table*}[h!]
  \centering
  \small
  \begin{tabular}{p{0.65\textwidth}|p{0.15\textwidth}}
    \toprule
    \textbf{Dialogue Context}: & \textbf{Dialogue Act}: \\
    ``Well , have you ever heard of Oulong tea and compressed tea ?'' &  ``directive'' \\
    ``Oh , yeah . Oulong tea is good for one's health . isn't it ?'' &  \\
    ``You surely know a lot about Chinese tea.'' &  \\
    ``Sure , I like drinking tea at teahouses.'' &  \\
    ``Oh , so do I.'' & \\
    \midrule
    \multicolumn{2}{l}{\textbf{Ground Truth:} ''Why don't we go for one now ?''} \\
    \multicolumn{2}{l}{\textbf{DialoGPT-Ori:} ''I'm sure.''} \\
    \multicolumn{2}{l}{\textbf{DialoGPT-FT:} ''I'm looking forward to tea together. Have you heard the news about that Chinese new food?''} \\
    \multicolumn{2}{l}{\textbf{DialoGPT-FUDGE:} ''I'm not sure if you're being serious or not, but I'm pretty sure that's not what he's saying.''} \\
    \multicolumn{2}{l}{\textbf{DialoGPT-PPLM:} ''It's an excellent way to clean yourself.''} \\
    \multicolumn{2}{l}{\textbf{DialoGPT-SideControl:} ''You should.''} \\
    \bottomrule
  \end{tabular}
  \caption{\label{tab:da-case-10k}
  Semantic label control generation results under 10000 training samples sampled from DailyDialog.
  }
\end{table*}

\begin{table*}[h!]
  \centering
  \small
  \begin{tabular}{p{0.65\textwidth}|p{0.15\textwidth}}
    \toprule
    \textbf{Dialogue Context}: & \textbf{Dialogue Act}: \\
    ``Well , have you ever heard of Oulong tea and compressed tea ?'' &  ``directive'' \\
    ``Oh , yeah . Oulong tea is good for one's health . isn't it ?'' &  \\
    ``You surely know a lot about Chinese tea.'' &  \\
    ``Sure , I like drinking tea at teahouses.'' &  \\
    ``Oh , so do I.'' & \\
    \midrule
    \multicolumn{2}{l}{\textbf{Ground Truth:} ''Why don't we go for one now ?''} \\
    \multicolumn{2}{l}{\textbf{DialoGPT-Ori:} ''I'm sure.''} \\
    \multicolumn{2}{l}{\textbf{DialoGPT-FT:} ''well, what kind of tea do you like?''} \\
    \multicolumn{2}{l}{\textbf{DialoGPT-FUDGE:} ''I'm not sure if you're being serious or not, but I'm pretty sure that's not what he's saying.''} \\
    \multicolumn{2}{l}{\textbf{DialoGPT-PPLM:} ''I know, but it's just tea.''} \\
    \multicolumn{2}{l}{\textbf{DialoGPT-SideControl:} ''I'm not in the mood to go.''} \\
    \bottomrule
  \end{tabular}
  \caption{\label{tab:da-case-full}
  Semantic label control generation results under full training samples from DailyDialog.
  }
\end{table*}

\end{document}